\documentclass[10pt,journal,compsoc,twoside]{IEEEtran}
\ifCLASSOPTIONcompsoc
  \usepackage[nocompress]{cite}
\else
  \usepackage{cite}
\fi
\ifCLASSINFOpdf
\else
\fi
\usepackage{epsfig}
\usepackage{graphicx}
\usepackage{amsmath}
\usepackage{amssymb}
\usepackage{dirtree}
\usepackage{subcaption}
\usepackage{booktabs}
\usepackage{wrapfig}
\begin{document}
%
\title{Learning Generalized Transformation Equivariant Representations via AutoEncoding Transformations}

%
%
%
%

\author{Guo-Jun~Qi,~\IEEEmembership{Senior Member,~IEEE},~Liheng Zhang,~Xiao Wang
\IEEEcompsocitemizethanks{\IEEEcompsocthanksitem G.-J. Qi was with the Futurewei Seattle Cloud Lab, Seattle, WA, 98006. \protect\\
E-mail: guojunq@gmail.com. More information at http://maple-lab.net.
\IEEEcompsocthanksitem L. Zhang was with the Department of Computer Science, University of Central Florida, Orlando, FL, 32816. \protect\\
E-mail: lihengzhang1993@knights.ucf.edu
\IEEEcompsocthanksitem X. Wang was with the Department of Computer Science, Purdue University, West Lafayette, IN, 47906. \protect\\
E-mail: wang3702@purdue.edu
}
}

%
%

\markboth{Qi et al. Transformations Are All You Need}%
{Qi et al. Learning Generalized Transformation Equivariant Representations via Autoencoding Transformations}
%



\IEEEtitleabstractindextext{%
\begin{abstract}
Transformation Equivariant Representations (TERs) aim to capture the intrinsic visual structures that equivary to various transformations by expanding the notion of {\em translation} equivariance underlying the success of Convolutional Neural Networks (CNNs).
For this purpose, we present both deterministic AutoEncoding Transformations (AET) and probabilistic AutoEncoding Variational Transformations (AVT) models to learn visual representations from generic groups of transformations.  While the AET is trained by directly decoding the transformations from the learned representations, the AVT is trained by maximizing the joint mutual information between the learned representation and transformations. This results in Generalized TERs (GTERs) equivariant against transformations in a more general fashion by capturing complex patterns of visual structures beyond the conventional linear equivariance under a transformation group. The presented approach can be extended to (semi-)supervised models by jointly maximizing the mutual information of the learned representation with both labels and transformations.
Experiments demonstrate the proposed models outperform the state-of-the-art models in both unsupervised and (semi-)supervised tasks.

\end{abstract}

}

\maketitle

\IEEEdisplaynontitleabstractindextext

%
\IEEEpeerreviewmaketitle

\IEEEraisesectionheading{\section{Introduction}\label{sec:introduction}}

\begin{figure}[t]
    \centering
    \begin{subfigure}[c]{0.33\textwidth}
        \includegraphics[width=\textwidth]{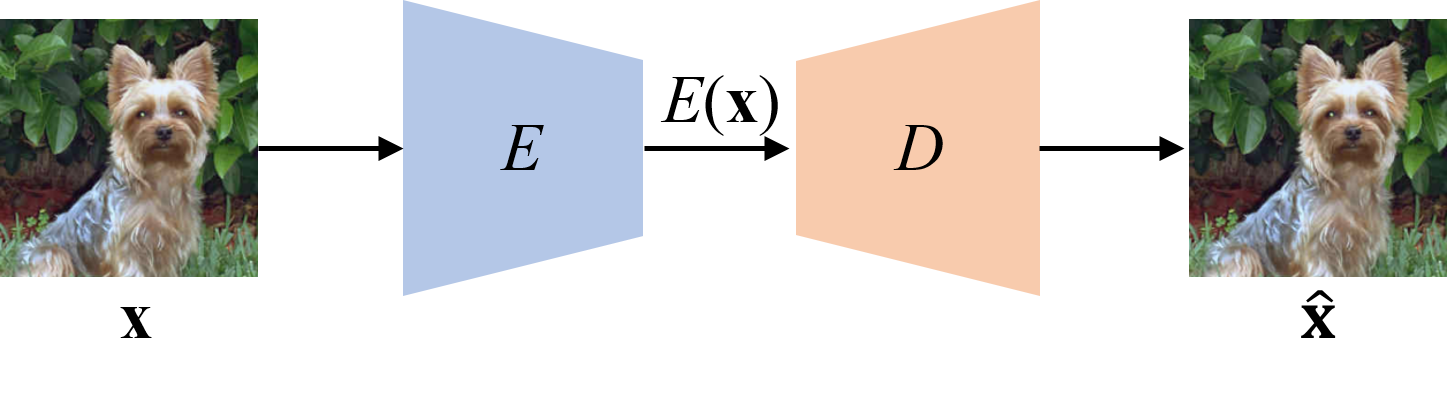}
        \caption{AutoEncoding Data (AED)}
    \end{subfigure}\\\vspace{2mm}
    ~ 
    \begin{subfigure}[c]{0.33\textwidth}
        \includegraphics[width=\textwidth]{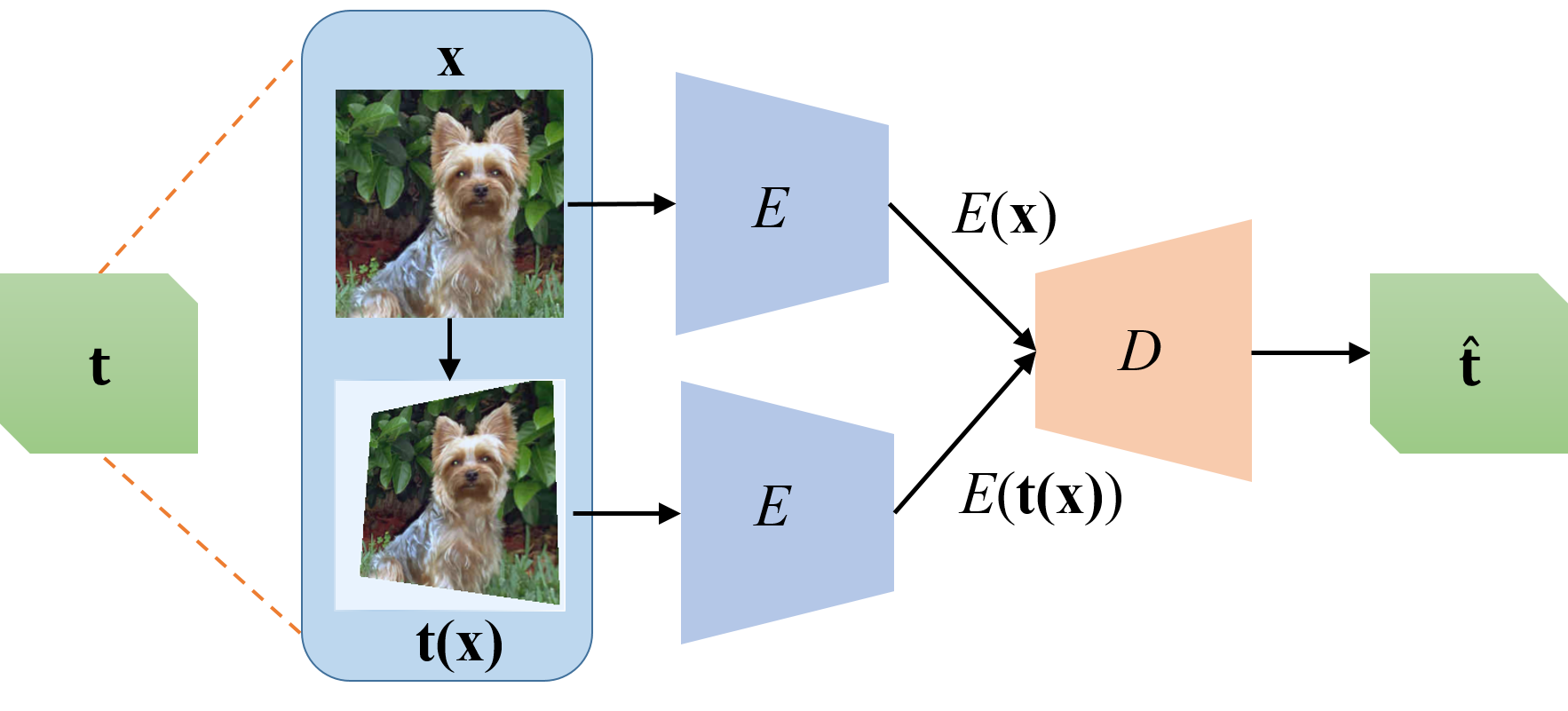}
        \caption{AutoEncoding Transformation (AET)}
    \end{subfigure}\\\vspace{2mm}
    \begin{subfigure}[c]{0.33\textwidth}
        \includegraphics[width=\textwidth]{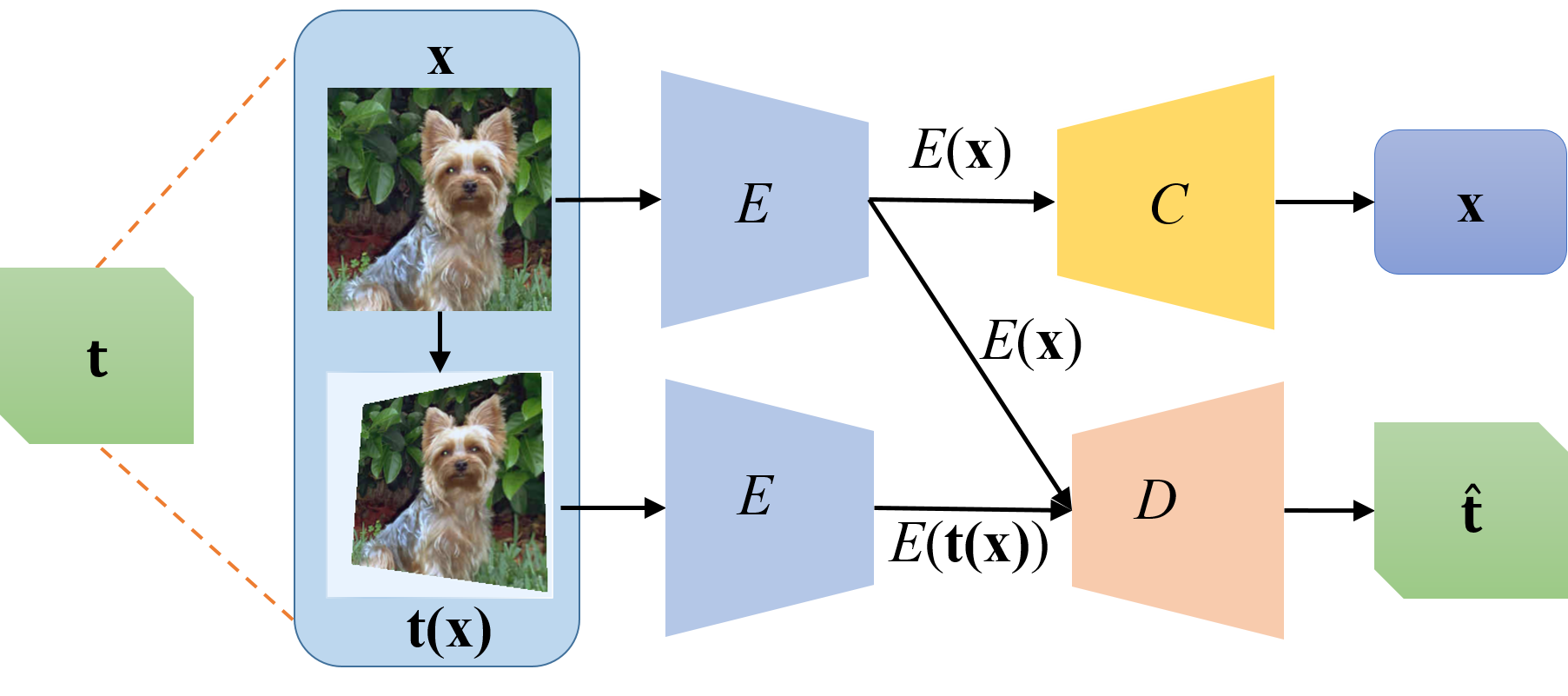}
        \caption{(Semi-)Supervised Autoencoding Transformation (SAT)}
    \end{subfigure}
    \caption{The figure illustrates a comparison between AED, AET, and AET. The AED and AET seek to reconstruct the input data and transformation at the output end, respectively. The encoder (E) extracts the representation of input and transformed images. The decoder (D) either reconstructs the data in AED, or the transformation in AET. The SAT builds a classifier (C) upon the output representation from the encoder by capturing the equivariant visual structures under various transformations. }\label{fig:comparison}
\end{figure}

%
%
%
%

\IEEEPARstart{I}n this paper, we aspire to show that {\em transformations} play a fundamental role in learning powerful representations by transforming images as a means to reveal the intrinsic patterns from transformed visual structures. Particularly,
Transformation Equivariant Representation (TER) learning seeks to model representations that equivary to various transformations on images.
In other words, the representation of an image ought to change in the same way as it is transformed. This is motivated by the assumption that image representations should capture the intrinsic visual structures such that transformations can be decoded from the representations of original and transformed images. Based on this assumption, we formally present a novel criterion of AutoEncoding Transformations (AET) to learn the TERs for various groups of transformations.

Learning the TERs has been adopted in Hiton's seminal work on learning transformation equivariant capsules \cite{hinton2011transforming}, and plays a critical role for the success of Convolutional Neural Networks (CNNs) \cite{krizhevsky2012imagenet}. Specifically, the representations learned by the CNNs are translation equivariant as their feature maps are shifted in the same way as input images are translated. On top of these feature maps that preserve the visual structures of translation equivariance, fully connected layers are built to output the predicted labels of input images.

Obviously, the translation equivariant convolutional features play the pivotal role in delivering the state-of-the-art performances in the deep networks. Thus, they are extended beyond translations to learn more expressive representations of equivariance to generic types of transformations, such as affine, projective and homographic transformations. Aline this direction, the group equivariant CNNs \cite{cohen2016group}
are developed to guarantee the transformation of input images results in the same transformation of input images.

However, the group equivariant CNNs \cite{cohen2016group} and their variants \cite{cohen2016steerable,lenssen2018group} are restricted to discrete transformations, and the resultant representations are also limited to a group representation of linear transformations. These limitations restrict their abilities to model group representations of complex transformations that could be continuous and nonlinear in many learning tasks, ranging from unsupervised, to semi-supervised and supervised learning.

\subsection{Unsupervised Learning of Transformation Equivariant Representations}

The focus of this paper is on the principle of autoencoding transformations and its application to learn the transformation equivariant representations. The core idea is to encode data with the representations from which the transformations can be decoded as much as possible. We will begin with an unsupervised learning of such representations without involving any labeled data, and then proceed to a generalization to semi-supervised and supervised representations by encoding label information as well.

Unlike group equivariant CNNs that learn the feature maps mathematically satisfying the transformation equivariance as a function of the group of transformations, the proposed AutoEncoding Transformations (AET) presents an autoencoding architecture to learn transformation equivariant representations by reconstructing applied transformations. As long as a transformation of input images results in equivariant representations, it should be well decoded from the representations of original and transformed images.  Compared with the group equivariant CNNS, the AET model is more flexible and tractable to tackle with any transformations and their compositions, since it does not rely on a strict convolutional structure to

The AET is also in contrast to the conventional AutoEncoding Data (AED) paradigm that instead aims to reconstruct data rather than the transformations. Figure \ref{fig:comparison}(a) and (b) illustrate the comparison between the AET and AED. Since the space of transformations (e.g., the few parameters of transformations) is of quite lower dimension than that of data space (e.g., the pixel space of images), the decoder of the AET can be quite shallower than that of the AED. This allows the backpropagated errors to more sufficiently train the encoder that models the representations of input data in the AET architecture.

Moreover, an AET model can be trained from an information-theoretic perspective by maximizing the information in the learned representation about the applied transformation and the input data. This will generalize the group representations of linear transformations to more general forms that could equivary nonlinearly to input transformations. It results in Generalized Transformation Equivariant Representations (GTERs) that can capture more complex patterns of visual structure under transformations. Unfortunately, this will result in an intractable optimization problem to maximize the mutual information between representations and transformations. A variational lower bound of the mutual information can be derive by introducing a surrogate transformation decoder, yielding a novel model of Autoencoding Variational Transformation (AVT) as an alterative to the deterministic AET.

\subsection{(Semi-)Supervised Learning of Transformation Equivariant Representations}

While both AET and AVT are trained in an unsupervised fashion, they can act as the basic representation for building the (semi-)supervised classifiers. Along this direction, we can train (Semi-)Supervised Autoencoding Transformation (SAT) that jointly trains the transformation equivariant representations as well as the corresponding classifiers.

Figure \ref{fig:comparison}(c) illustrates the SAT model, where a classifier head is added upon the representation encoder of an AET network.  The SAT can be based on either the deterministic AET or the probabilistic AVT architecture. Particularly, along the direction pointed by the AVT, we seek to train the proposed (semi-)supervised transformation equivariant classifiers by maximizing the mutual information of the learned representations with the transformations and labels. In this way, the trained SAT model can not only handle the transformed data through their equivarying representations, but also encode the labeling information through the supervised classifier. The resultant SAT also contains the deterministic model based on the AET as a special case by fixing a deterministic model to representation encoder and the transformation decoder.

The transformation equivariance in the SAT model is contrary to the data augmentation by transformations in deep learning literature \cite{krizhevsky2012imagenet}. First, the data augmentation is only applicable to augment the labeled examples for model training, which cannot be extended to unlabeled data. This limits it in semi-supervised learning by exploring the unlabeled data. Second, the data augmentation aims to enforce the transformation invariance, in which the labels of transformed data are supposed to be invariant. This differs from the motivation to encode the inherent visual structures that equivary under various transformations.

Actually, in the (semi-)supervised transformation equivariant classifiers, we aim to integrate the principles of both training {\em transformation equivariant representations} and {\em transformation invariant classifiers} seamlessly. Indeed, both principles have played the key role in compelling performances of the CNNs and their modern variants. This is witnessed by the translation equivariant convolutional feature maps and the atop classifiers that are supposed to make transformation-invariant predictions with the spatial pooling and fully connected layers. We will show that the proposed SAT extends the translation equivariance in the CNNs to cover a generic class of transformation equivariance, as well as encode the labels to train the representations and the associated transformation invariant classifiers. We hope this can deepen our understanding of the interplay between the transformation equivariance and invariance both of which play the fundamental roles in training robust classifiers with labeled and unlabeled data.

The remainder of this paper is organized as follows. We will review the related works in Section~\ref{sec:related}.  The unsupervised and (semi-)supervised learning of transformation equivariant representations will be presented in the autoencoding transformation framework in Section~\ref{sec:unsup} and Section~\ref{sec:sup}, respectively.  We will present experiment results in Section~\ref{sec:unexp} and Section~\ref{sec:supexp} for unsupervised and semi-supervised tasks. We will conclude the paper and discuss the future works in Section~\ref{sec:concl}.

\section{Related Works}\label{sec:related}
In this section, we will review the related works on learning transformation-equivariant representation, as well as unsupervised and (semi-)supervised models.

\subsection{Transformation-Equivariant Representations}
Learning transformation-equivariant representations can trace back to the seminal work on training capsule nets \cite{sabour2017dynamic,hinton2011transforming,hinton2018matrix}. The transformation equivariance is characterized by the various directions of capsules, while the confidence of belonging to a particular class is captured by their lengths.

Many efforts have been made in literature  \cite{cohen2016group,cohen2016steerable,lenssen2018group} on extending the conventional translation-equivariant convolutions to cover more transformations.
Among them are group equivariant convolutions (G-convolution) \cite{cohen2016group} that have been developed to equivary to more types of transformations. The idea of group equivariance has also been introduced to the capsule nets \cite{lenssen2018group} by ensuring the equivariance of output pose vectors to a group of transformations with a generic routing mechanism. However, the group equivariant convolution is restricted to discrete transformations, which limits its ability to learn the representations equivariant to generic continuous transformations.


%
%

\subsection{Unsupervised Representation Learning}
{\noindent \bf Auto-Encoders and GANs.}
Unsupervised auto-encoders have been extensively studied in literature \cite{hinton1994autoencoders,japkowicz2000nonlinear,vincent2008extracting}. Existing auto-encoders are trained by reconstructing input {\em data} from the outputs of encoders.
A large category of auto-encoder variants have been proposed.
Among them is the Variational Auto-Encoder (VAE) \cite{kingma2013auto} that maximizes the lower-bound of the data likelihood to train a pair of probabilistic encoder and decoder, while beta-VAE seeks to disentangle representations by introducing an adjustable hyperparameter on the capacity of latent channel to balance between the independence constraint and the reconstruction accuracy \cite{higgins2017beta}.  Denoising auto-encoders \cite{vincent2008extracting}
attempt to reconstruct noise-corrupted data to
learn robust representations, while
contrastive Auto-Encoders \cite{rifai2011contractive} encourage to learn representations invariant to small perturbations on data.
Along this direction, Hinton et al.~\cite{hinton2011transforming} propose capsule networks to explore transformation equivariance by minimizing the discrepancy between the reconstructed and target data.

On the other hand, Generative Adversarial Nets (GANs) have also been used to train unsupervised representations.  Unlike the auto-encoders, the GANs \cite{goodfellow2014generative} and their variants \cite{donahue2016adversarial,dumoulin2016adversarially,qi2017loss,arjovsky2017wasserstein} generate data from the noises drawn from a simple distribution, with a discriminator trained adversarially to distinguish between real and fake data. The sampled noises can be viewed as the representation of generated data over a manifold, and one can train an encoder by inverting the generator to find the generating noise. This can be implemented by jointly training a pair of mutually inverse generator and encoder \cite{donahue2016adversarial,dumoulin2016adversarially}. There also exist better generalizable GANs in producing unseen data based on the Lipschitz assumption on the real data distribution \cite{qi2017loss,arjovsky2017wasserstein}, which can give rise to more powerful representations of data out of training examples  \cite{donahue2016adversarial,dumoulin2016adversarially,edraki2018generalized}. Compared with the Auto-Encoders, GANs do not rely on learning one-to-one reconstruction of data; instead, they aim to generate the entire distribution of data.

{\noindent \bf Self-Supervisory Signals.} There exist many other unsupervised learning methods using different types of self-supervised signals to train deep networks.
Mehdi and Favaro~\cite{noroozi2016unsupervised} propose to solve Jigsaw puzzles to train a convolutional neural network.
Doersch et al.~\cite{doersch2015unsupervised} train the network by inferring the relative positions between sampled patches from an image as self-supervised information. Instead, Noroozi et al.~\cite{noroozi2017representation} count features that satisfy equivalence relations between downsampled and tiled images. Gidaris et al.~\cite{gidaris2018unsupervised} propose to train RotNets by predicting a discrete set of image rotations, but they are unable to handle generic continuous transformations and their compositions. Dosovitskiy et al.~\cite{dosovitskiy2014discriminative} create a set of surrogate classes by applying various transformations to individual images. However, the resultant features could over-discriminate visually similar images as they always belong to different surrogate classes.
Unsupervised features have also been learned from videos by estimating the self-motion of moving objects between consecutive frames \cite{agrawal2015learning}.

\subsection{(Semi-)Supervised Representation Learning}

In addition, there exist a large number of semi-supervised models in literature. Here, we particularly mention three  state-of-the-art methods that will be compared in experiments.  Temporal ensembling \cite{laine2016temporal} and mean teachers \cite{tarvainen2017mean} both use an ensemble of teachers to supervise the training of a student model. Temporal ensembling uses the exponential moving average of predictions made by past models on unlabeled data as targets to train the student model. Instead, mean teachers update the student model with the exponential moving average of the weights of past models. On the contrary, the Virtual Adversarial Training (VAT) \cite{miyato2018virtual} seeks to minimizes the change of predictions on unlabeled examples when their output values are adversarially altered. This could result in a robust model that prefers smooth predictions over unlabeled data.

The SAT also differs from transformation-based data augmentation in which the transformed samples and their labels are used directly as additional training examples \cite{krizhevsky2012imagenet}. First, in the semi-supervised learning, unlabeled examples cannot be directly augmented to form training examples due to their missing labels. Moreover, data augmentation needs to preserve the labels on augmented images, and this prevents us from applying the transformations that could severely distort the images (e.g., shearing, rotations with arbitrary angles, and projective transformations) or invalidate the associated labels (e.g., vertically flipping ``6" to ``9").
In contrast, the SAT avoids using the labels of transformed images to supervisedly train the classifier directly; instead it attempts to encode the visual structures of images equivariant to various transformations without access to their labels. This leads to a label-blind TER regularizer to explore the unlabeled examples for the semi-supervised problem.

\section{Unsupervised Learning of Transformation Equivariant Representations}\label{sec:unsup}

In this section, we will first present the autoencoding transformation architecture to learn the transformation equivariant representations in a deterministic fashion. Then, a variational alternative approach will be presented to handle the uncertainty in the representation learning by maximizing the mutual information between the learned representations and the applied transformations.

\subsection{AET: A Deterministic Model}

We begin by defining the notations used in the proposed AutoEncoding Transformation (AET) architecture.
Consider a random transformation $\mathbf t$ sampled from a transformation distribution $p(\mathbf t)$ (e.g., warping, projective and homographic transformations), as well as an image $\mathbf x$ drawn from a data distribution $p(\mathbf x)$ in a sample space $\mathcal X$. Then the application of $\mathbf t$ to $\mathbf x$ results in a transformed image $\mathbf t(\mathbf x)$.

The goal of AET focuses on learning a representation encoder $E_\theta: \mathbf x \mapsto E_\theta(\mathbf x)$ with parameters $\theta$, which maps a sample $\mathbf x\sim p(\mathbf x)$ to its representation $E_\theta(\mathbf x)$ in a linear space $\mathcal Z$.
For this purpose, one need to learn a transformation decoder with parameters $\phi$
$$
D_\phi: \left[E_\theta(\mathbf x), E_\theta(\mathbf t(\mathbf x))\right]\mapsto \hat {\mathbf t}
$$
that makes an estimate $\hat{\mathbf t}$ of the input transformation $\mathbf t$ from the representations of original and transformed samples. Since the transformation decoder takes the encoder outputs rather than original and transformed images, this pushes the encoder to capture the inherent visual structures of images to make a satisfactory estimate of the transformation.

Then the AET can be trained to jointly learn the representation encoder $E_\theta$ and the transformation decoder $D_\phi$. A loss function $\ell(\mathbf t, \hat{\mathbf t})$ measuring the deviation between a transformation $\mathbf t$ and its estimate $\hat{\mathbf t}$ is minimized to train the AET over $p(\mathbf t)$ and $p(\mathbf x)$:
\begin{equation}\label{eq:AET_loss}
\min_{\theta,\phi} \mathop\mathbb E\limits_{\mathbf t \sim p(\mathbf t), \mathbf x\sim p(\mathbf x)}\ell(\mathbf t, \hat{\mathbf t})
\end{equation}
where the estimated transformation $\hat{\mathbf t}$ can be written as a function of the encoder $E_\theta$ and the decoder $D_\phi$ such that
$$
\hat{\mathbf t}= D_\phi\left[E_\theta(\mathbf x), E_\theta(\mathbf t(\mathbf x))\right],
$$
and the expectation $\mathbb E$ is taken over the distributions of transformations and data.

In this way, the encoder $E_\theta$ and the decoder $D_\phi$ can be jointly trained over mini-batches by back-propagating the gradient of the loss $\ell$ to update their parameters.

\subsection{AVT: A Probabilistic Model}

Alternatively, we can train transformation equivariant representations to contain as much information as possible about applied transformations to recover them.

\begin{figure*}[t]
    \centering
    \begin{subfigure}[c]{0.49\textwidth}
        \includegraphics[width=\textwidth]{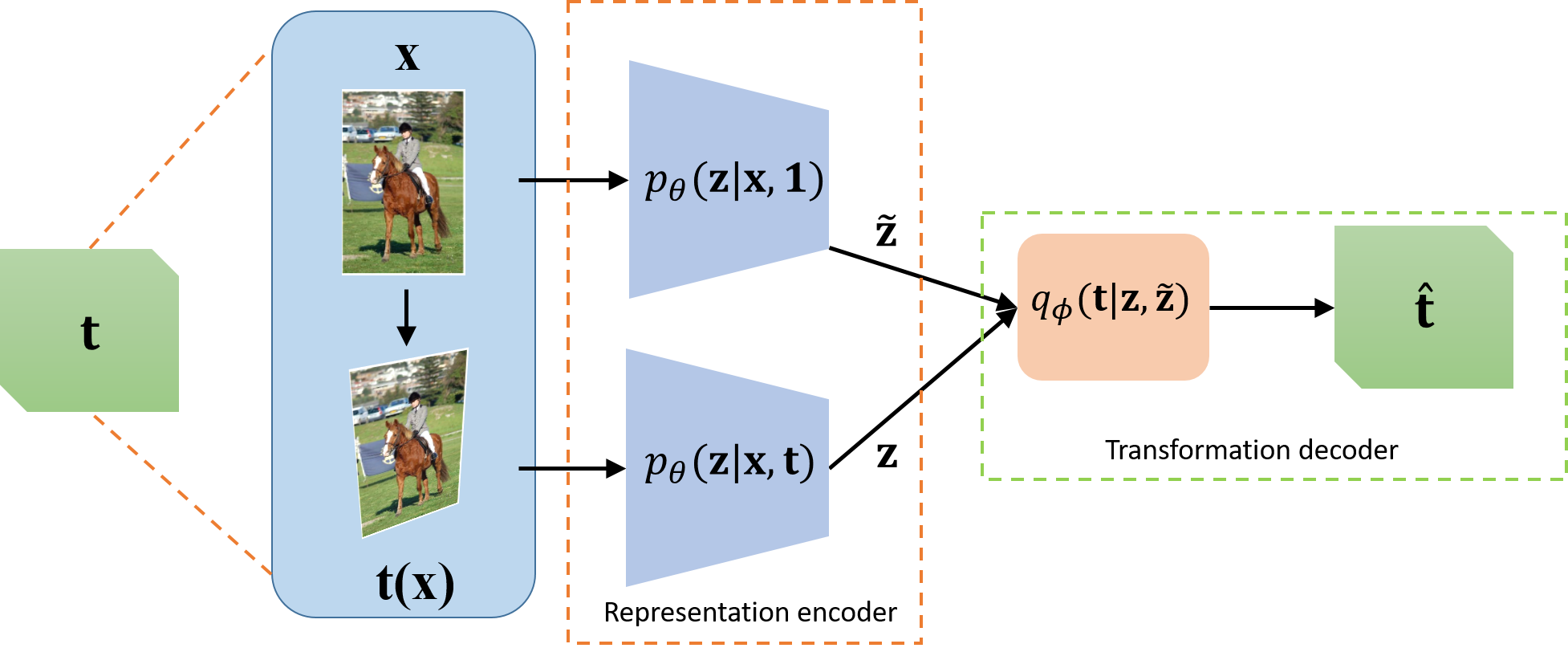}
        \caption{Autoencoding Variational Transformations (AVT)}
    \end{subfigure}
    ~ 
    \begin{subfigure}[c]{0.49\textwidth}
        \includegraphics[width=\textwidth]{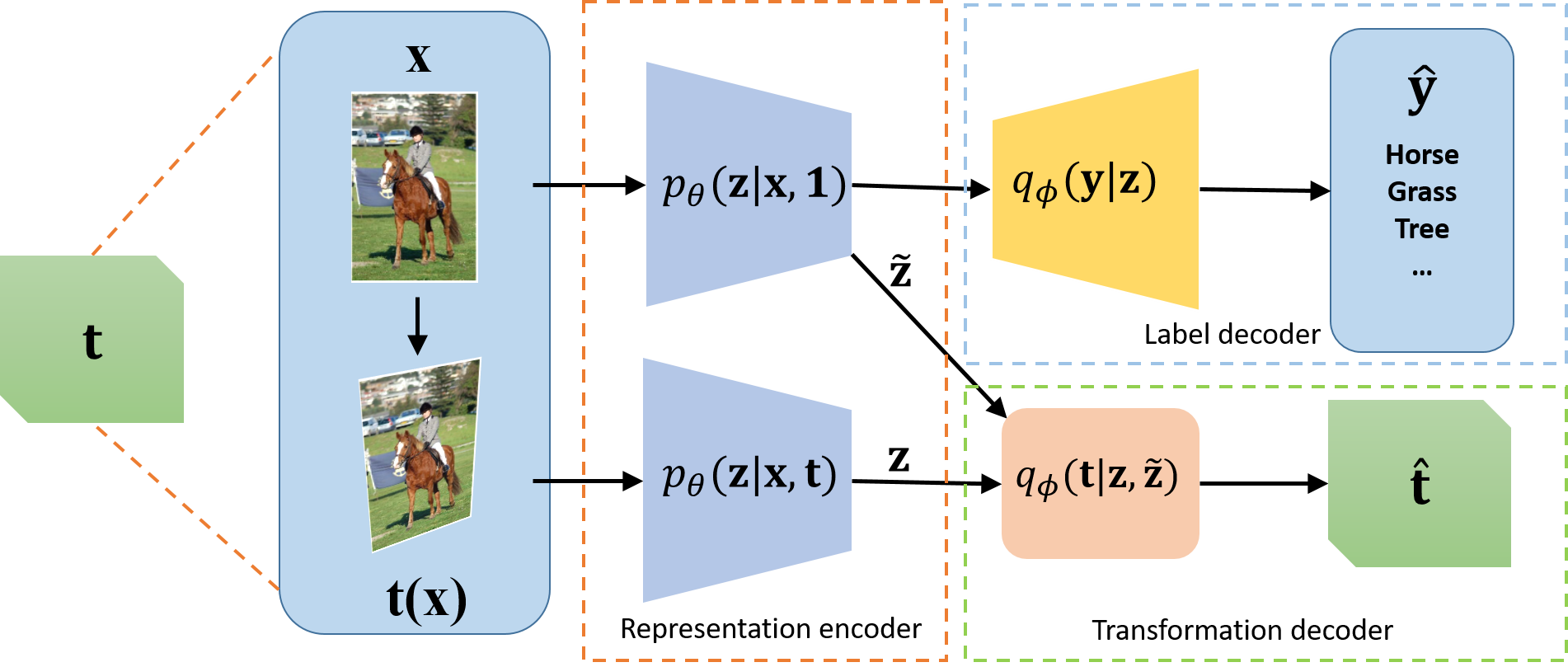}
        \caption{(Semi-)Supervised Autoencoding Transformations (SAT)}
    \end{subfigure}\\
    \caption{The figure illustrates the variational approach to unsupervised learning and (semi-)supervised learning of autoencoding transformations, namely AVT and SAT respectively. The probability $p_\theta(\mathbf z|\mathbf t,\mathbf x)$ acts as the representation encoder, while $q_\phi(\mathbf t|\mathbf z,\mathbf {\tilde z})$ and $q_\phi(\mathbf y|\mathbf {\tilde z})$ play the roles of a transformation and label decoder, respectively. By setting the transformation to an identity $\mathbf 1$, the corresponding $\tilde{\mathbf z}$ is the representation of an original image.}\label{fig:variational}
\end{figure*}

\subsubsection{Notations}
Formally, our goal is to learn an encoder that maps a transformed sample $\mathbf t(\mathbf x)$ to a probabilistic representation with the mean $f_\theta$ and variance $\sigma_\theta$. This results in the following probabilistic representation $\mathbf z \in \mathcal Z$ of $\mathbf t(\mathbf x)$:
\begin{equation}\label{eq:rep_t}
\mathbf z = f_\theta(\mathbf t(\mathbf x))+\sigma_\theta(\mathbf t(\mathbf x)) \circ \epsilon
\end{equation}
where $\epsilon$ is sampled from a normal distribution $p(\epsilon)\triangleq\mathcal N(\epsilon|\mathbf 0, \mathbf I)$ with $\circ$ denoting the element-wise product. Thus, the resultant probabilistic representation $\mathbf z$ follows a normal distribution
$$
p_\theta(\mathbf z|\mathbf t, \mathbf x)\triangleq \mathcal N\left(\mathbf z|f_\theta(\mathbf t(\mathbf x)),\sigma_\theta^2(\mathbf t(\mathbf x))\right)
$$
conditioned on the randomly sampled transformation $\mathbf t$ and input data $\mathbf x$.

On the other hand, the representation of the original sample $\mathbf x$ is a special case when $\mathbf t$ is an identity transformation, which is
\begin{equation}\label{eq:rep_ori}
\mathbf {\tilde z} = f_\theta(\mathbf x)+\sigma_\theta(\mathbf x) \circ \tilde \epsilon
\end{equation}
whose mean and variance are computed by using the deep network with the same weights $\theta$, and $\tilde \epsilon \sim p(\tilde\epsilon)\triangleq\mathcal N(\tilde\epsilon|\mathbf 0, \mathbf I)$.

\subsubsection{Generalized Transformation Equivariance}
In the conventional definition of transformation equivariance, there should exist an automorphism $\boldsymbol\rho(\mathbf t)\in {\rm Aut}(\mathcal Z):\mathcal Z \rightarrow \mathcal Z$ in the representation space, such that \footnote{The transformation $\mathbf t$ in the sample space $\mathcal X$ and the corresponding transformation $\boldsymbol\rho$ in the representation space $\mathcal Z$ need not be the same. But the representation transformation $\boldsymbol\rho(\mathbf t)$ should be a function of the sample transformation $\mathbf t$. }
$$
\mathbf z = [\boldsymbol\rho (\mathbf t)](\mathbf {\tilde z})
$$
Here the transformation $\boldsymbol\rho(\mathbf t)$ is independent of the input sample $\mathbf x$. In other words, the representation $\mathbf z$ of a transformed sample is completely determined by the original representation $\mathbf {\tilde z}$ and the applied transformation $\mathbf t$ with no need to access the sample $\mathbf x$. This is called {\em steerability} property in literature \cite{cohen2016steerable}, which enables us to compute  $\mathbf z$ by applying the sample-independent transformation directly to the original representation $\mathbf {\tilde z}$.

This property can be generalized without relying on the linear group representations of transformations through automorphisms. Instead of sticking with a linear $\boldsymbol\rho(\mathbf t)$, one can seek a more general relation between $\mathbf z$ and $\mathbf {\tilde z}$, independently of $\mathbf x$.  From an information theoretical point of view, this requires $(\mathbf {\tilde z},\mathbf t)$ should jointly contain all necessary information about $\mathbf z$ so that $\mathbf z$ can be best estimated from them without a direct access to $\mathbf x$.


This leads us to maximizing the mutual information $I_\theta(\mathbf z;\mathbf {\tilde z},\mathbf t)$ to learn the generalized transformation equivariant representations. Indeed, by the chain rule and the nonnegativity of mutual information, we have
$$
I_\theta(\mathbf z;\mathbf {\tilde z},\mathbf t)= I_\theta(\mathbf z;\mathbf {\tilde z},\mathbf t, \mathbf x) - I_\theta(\mathbf z;\mathbf x|\mathbf {\tilde z},\mathbf t) \leq I_\theta(\mathbf z;\mathbf {\tilde z},\mathbf t, \mathbf x),
$$
which shows $I_\theta(\mathbf z;\mathbf {\tilde z},\mathbf t)$ is upper bounded by the mutual information $I_\theta(\mathbf z;\mathbf {\tilde z},\mathbf t, \mathbf x)$ between $\mathbf z$ and $(\mathbf {\tilde z},\mathbf t,\mathbf x)$.

Clearly, when $I_\theta(\mathbf z;\mathbf x|\mathbf {\tilde z},\mathbf t)=0$, $I_\theta(\mathbf z;\mathbf {\tilde z},\mathbf t)$ attains the maximum value of its upper bound $I_\theta(\mathbf z;\mathbf {\tilde z},\mathbf t, \mathbf x)$. In this case, $\mathbf x$ would provide no more information about $\mathbf z$ than $(\mathbf {\tilde z},\mathbf t)$, which implies one can estimate $\mathbf z$ directly from $(\mathbf {\tilde z},\mathbf t)$ without accessing $\mathbf x$.
Thus, we propose to solve
$$
\theta^\star = \arg\max_{\theta} I_\theta(\mathbf z;\mathbf {\tilde z},\mathbf t)
$$
to learn the probabilistic encoder $\theta^\star$ in pursuit of such a generalized TER.

However, a direction maximization of the above mutual information needs to evaluate an intractable posterior $p_\theta(\mathbf t|\mathbf z,\mathbf {\tilde z})$ of the transformation. Thus, we instead lower bound the mutual information by introducing a surrogate decoder $q_\phi(\mathbf t|\mathbf z,\mathbf {\tilde z})$ with the parameters $\phi$ to approximate the true posterior.

\subsubsection{Variational Approach}

Unlike the variational autoencoder that lower-bounds data likelihood \cite{kingma2013auto}, we directly take a lower bound of the mutual information \cite{agakov2004algorithm} between  $\mathbf z$ and $(\mathbf {\tilde z},\mathbf t)$ below
\[
\begin{aligned}
&I_\theta(\mathbf z; \mathbf {\tilde z},\mathbf t) = I_\theta(\mathbf z; \mathbf {\tilde z}) + I_\theta(\mathbf z; \mathbf t|\mathbf {\tilde z})\\
&\geq I_\theta(\mathbf z; \mathbf t|\mathbf {\tilde z})
= H(\mathbf t|\mathbf {\tilde z}) - H(\mathbf t|\mathbf z,\mathbf {\tilde z})\\
&=H(\mathbf t|\mathbf {\tilde z}) +\mathop\mathbb E\limits_{p_\theta(\mathbf t,\mathbf z,\mathbf {\tilde z})} \log p_\theta(\mathbf t|\mathbf z,\mathbf {\tilde z})\\
&= H(\mathbf t|\mathbf {\tilde z})  + \mathop\mathbb E\limits_{p_\theta(\mathbf t,\mathbf z,\mathbf {\tilde z})} \log q_\phi(\mathbf t|\mathbf z,\mathbf {\tilde z})\\
&+ \mathop\mathbb E\limits_{p(\mathbf z,\mathbf {\tilde z})} D(p_\theta(\mathbf t|\mathbf z,\mathbf {\tilde z})\|q_\phi(\mathbf t|\mathbf z,\mathbf {\tilde z}))\\
&\geq H(\mathbf t|\mathbf {\tilde z})  + \mathop\mathbb E\limits_{p_\theta(\mathbf t,\mathbf z,\mathbf {\tilde z})} \log q_\phi(\mathbf t|\mathbf z,\mathbf {\tilde z}) \triangleq \tilde I_{\theta,\phi}(\mathbf z;\mathbf {\tilde z},\mathbf t)
\end{aligned}
\]
where $H(\cdot)$ denotes the (conditional) entropy, and $D(p_\theta(\mathbf t|\mathbf z,\mathbf {\tilde z})\|q_\phi(\mathbf t|\mathbf z,\mathbf {\tilde z}))$ is the non-negative Kullback divergence between $p_\theta$ and $q_\phi$.

We choose to maximize the lower variational bound $\tilde I_{\theta,\phi}(\mathbf z;\mathbf {\tilde z},\mathbf t)$. Since $H(\mathbf t|\mathbf {\tilde z})$ is nonnegative and independent of the model parameters $\theta$ and $\phi$, we choose to solve
\begin{equation}\label{eq:var}
\begin{aligned}
&\max\limits_{\theta,\phi}~\mathcal L^{\rm unsup}_{\theta,\phi} \triangleq \mathop\mathbb E\limits_{p_\theta(\mathbf t,\mathbf z,\mathbf {\tilde z})}\log q_\phi(\mathbf t|\mathbf z,\mathbf {\tilde z})\\
&=\mathop\mathbb E\limits_{p(\mathbf x),p(\mathbf t)} \mathop\mathbb E\limits_{p(\epsilon),p(\tilde\epsilon)}\log q_\phi(\mathbf z,\mathbf {\tilde z})\\
\end{aligned}
\end{equation}
to learn $\theta$ and $\phi$ under the expectation over $p(\mathbf t, \mathbf z, \mathbf {\tilde z})$, and the equality follows from the generative process for the representations in Eqs.~(\ref{eq:rep_t})--(\ref{eq:rep_ori}).

\subsubsection{Variational Transformation Decoder}\label{sec:var_decoder}

To estimate a family of continuous transformations, we choose a normal distribution $\mathcal N(\mathbf t|d_\phi(\mathbf z,\mathbf {\tilde z}),\sigma^2_\phi(\mathbf z,\mathbf {\tilde z}))$ as the posterior $q_\phi(\mathbf t|\mathbf z,\mathbf {\tilde z})$ of the transformation decoder, where the mean $d_\phi(\mathbf z,\mathbf {\tilde z})$ and variance $\sigma^2_\phi(\mathbf z, \mathbf {\tilde z})$ are implemented by deep network respectively.

For categorical transformations (e.g., horizontal vs. vertical flips, and rotations of different directions), a categorical distribution ${\rm Cat}(\mathbf t|\boldsymbol\pi_\phi(\mathbf z,\mathbf {\tilde z}))$ can be adopted as the posterior $q_\phi(\mathbf t|\mathbf z,\mathbf {\tilde z})$, where each entry of $\boldsymbol\pi_\phi(\mathbf z,\mathbf {\tilde z})$ is the probability mass for a transformation type. A hybrid distribution can also be defined to combine multiple continuous and categorical transformations, making the variational transformation decoder more flexible and appealing in handling complex transformations.

The posterior $q_\phi(\mathbf t|\mathbf z,\mathbf {\tilde z})$ of transformation is a function of the representations of the original and transformed images. Thus, a natural choice is to use a Siamese encoder network with shared weights to output the representations of original and transformed samples, and construct the transformation decoder atop the concatenated representations. Figure \ref{fig:variational}(a) illustrates the architecture of the AVT network.

Finally, it is not hard to see that the deterministic AET model would be viewed as a special case of the AVT, if the probabilistic representation encoder $p_\theta(\mathbf z|\mathbf t,\mathbf x)$ and transformation decoder $q_\phi(\mathbf t|\mathbf z,\mathbf {\tilde z})$ were set to deterministic forms as in the AET.


\section{(Semi-)Supervised Learning of Transformation Equivariant Representations}\label{sec:sup}

Autoencoding transformations can act as the basic representation block in many learning problems. In this section, we present its role in (semi-)supervised learning tasks to enable more accurate classification of samples by capturing their transformation equivariant representations.

\subsection{SAT: (Semi-)Supervised Autoencoding Transformations}

The unsupervised learning of autoencoding transformations can be generalized to (semi-)supervised cases with labeled samples. Accordingly, the goal is formulated as learning of representations that contain as much (mutual) information as possible about not only applied transformations but also data labels.

Given a labeled sample $(\mathbf x,\mathbf y)$, we can define the joint distribution over the representation, transformation and label,
$$
p_\theta(\mathbf y, \mathbf t, \mathbf z, \mathbf {\tilde z}|\mathbf x) =  p(\mathbf t) p_\theta(\mathbf {\tilde z}|\mathbf x) p_\theta(\mathbf z | \mathbf t, \mathbf x) p(\mathbf y|\mathbf x)
$$
where we have assumed that $\mathbf y$ is independent of $\mathbf t$ and $\mathbf z$ once the sample $\mathbf x$ is given.

In presence of sample labels, the pursuit of transformation equivariant representations can be performed by maximizing the joint mutual information $I_\theta(\mathbf y, \mathbf z; \mathbf t, \mathbf {\tilde z})$ such that the representation $\mathbf z$ of the original sample and the transformation $\mathbf t$  contains sufficient information to classify the label $\mathbf y$ as well as learn the representation $\mathbf z$ equivariant to the transformed sample.

Like in (\ref{eq:var}) for the unsupervised case, the joint mutual information can be lower bounded in the following way,
\[
\begin{aligned}
&I_\theta(\mathbf y,\mathbf z;\mathbf{\tilde z},\mathbf t) = I_\theta(\mathbf y,\mathbf z;\mathbf{\tilde z}) + I_\theta(\mathbf y,\mathbf z; \mathbf t|\mathbf{\tilde z})\\
&=(I_\theta(\mathbf z;\mathbf{\tilde z}) + I_\theta(\mathbf y,\mathbf{\tilde z} | \mathbf z))+ (I_\theta(\mathbf z;\mathbf t|\mathbf{\tilde z}) + I_\theta(\mathbf y;\mathbf t|\mathbf z,\mathbf{\tilde z}))\\
&\geq I_\theta(\mathbf y,\mathbf{\tilde z} | \mathbf z)+I_\theta(\mathbf z;\mathbf t|\mathbf{\tilde z})\\
&\geq H(\mathbf y|\mathbf z) + \mathop\mathbb E\limits_{p_\theta(\mathbf y,\mathbf z,\mathbf {\tilde z})} \log q_\phi(\mathbf y|\mathbf z,\mathbf {\tilde z})\\
&+H(\mathbf t|\mathbf {\tilde z}) + \mathop\mathbb E\limits_{p_\theta(\mathbf t,\mathbf z,\mathbf {\tilde z})} \log q_\phi(\mathbf t|\mathbf z,\mathbf {\tilde z})\\
&\triangleq \tilde I_{\theta,\phi}(\mathbf y,\mathbf z;\mathbf{\tilde z},\mathbf t)
\end{aligned}
\]
where the first two equalities apply the chain rule of mutual information, and the first inequality uses the nonnegativity of the mutual information. In particular, we usually have $I_\theta(\mathbf y;\mathbf t|\mathbf z,\mathbf{\tilde z})=0$, which means the transformation should not change the label $\mathbf y$ of a sample (i.e., transformation invariance of sample labels). The second inequality follows the variational bound we derived earlier in the last section.

One can also assume the surrogate posterior $q_\phi(\mathbf y|\mathbf z,\mathbf {\tilde z})$ of labels can be simplified to $q_\phi(\mathbf y|\mathbf {\tilde z})$ since the representation of the original sample is supposed to provide sufficient information to predict the label.

Since $H(\mathbf y|\mathbf z)\geq 0$ and $H(\mathbf y,\mathbf t|\mathbf x)$ is independent of the model parameters $\theta$ and $\phi$, we maximize the following variational lower bound
\begin{equation}\label{eq:lb}
\begin{aligned}
&\max\limits_{\theta,\phi}
\mathcal L^{\rm sup}_{\theta,\phi}\triangleq
 \mathop\mathbb E\limits_{p_\theta(\mathbf y,\mathbf {\tilde z})} \log q_\phi(\mathbf y|\mathbf {\tilde z})+\mathop\mathbb E\limits_{p_\theta(\mathbf t,\mathbf z,\mathbf {\tilde z})} \log q_\phi(\mathbf t|\mathbf z,\mathbf {\tilde z})\\
 &=\mathop\mathbb E\limits_{p(\mathbf x)}\left\{\mathop\mathbb E\limits_{p(\mathbf y|\mathbf x),p(\tilde\epsilon)} \log q_\phi(\mathbf y|\mathbf {\tilde z})+\mathop\mathbb E\limits_{p(\mathbf t),p(\epsilon),p(\tilde\epsilon)} \log q_\phi(\mathbf t|\mathbf z, \mathbf {\tilde z})\right\}
\end{aligned}
\end{equation}
where $\mathbf z$ and $\mathbf {\tilde z}$ are sampled by following Eqs.~(\ref{eq:rep_t})--(\ref{eq:rep_ori}) in the equality, and the ground truth $\mathbf y$ is sampled from the label distribution $p(\mathbf y|\mathbf x)$ directly.

In a deterministic case, it is not hard to show that the first term of (\ref{eq:lb}) is related to the cross-entropy loss in training a supervised classifier, while the second term would reduce to the loss (\ref{eq:AET_loss}) in the deterministic AET model. Therefore, in this sense, the AET loss plays a role to regularize the  cross-entropy loss to train a supervised model.

In addition, a semi-supervised model can be trained by combining the unsupervised and supervised objectives (\ref{eq:var}) and (\ref{eq:lb})
\begin{equation}\label{eq:semi}
\max\limits_{\theta,\phi}
\mathcal L^{\rm unsup}_{\theta,\phi} + \lambda~\mathcal L^{\rm sup}_{\theta,\phi}
\end{equation}
with a nonnegative balancing coefficient $\lambda$. This enables to jointly explore labeled and unlabeled examples and their representations equivariant to various transformations.

We will demonstrate that the SAT can achieve superior performances to the existing state-of-the-art (semi-)supervised models. Moreover, the competitive performances also show great potentials of the model as the basic representation block in many machine learning and computer vision tasks.

Figure~\ref{fig:variational}(b) illustrates the architecture of the SAT model, in a comparison with its AVT counterpart. Particularly, in the SAT, the transformation and label decoders are jointly trained atop the representation encoder.

%
%

%

\section{Experiments: Unsupervised Learning}\label{sec:unexp}
In this section, we compare the proposed deterministic AET and probabilistic AVT models against the other unsupervised methods on the CIFAR-10, ImageNet and Places datasets.  The evaluation follows the protocols widely adopted by many existing unsupervised methods by applying the learned representations to downstream tasks.

\subsection{CIFAR-10 Experiments}
First, we evaluate the AET and AVT models on the CIFAR-10 dataset.

\subsubsection{Experiment Settings}


\vspace{2mm}
{\noindent \bf Architecture} To make a fair and direct comparison with existing models, the Network-In-Network (NIN) is adopted on the CIFAR-10 dataset for the unsupervised learning task \cite{gidaris2018unsupervised,zhang2019aet}.  
The NIN consists of four convolutional blocks, each of which contains three convolutional layers.
Both AET and AVT have two NIN branches with shared weights, each taking the original and transformed images as its input, respectively. The output features of the forth block of two branches are concatenated and average-pooled to form a $384$-d feature vector. Then an output layer follows to output the predicted transformation for the AET, and the mean $d_\phi$ and the log-of-variance $\log \sigma_\phi^2$ of the predicted transformation for the AVT, with the logarithm scaling the variance to a real value.

The first two blocks of each branch are used as the encoder network to output the deterministic representation for the AET, and the mean $f_\theta$ of the probabilistic representation for the AVT. An additional $1\times 1$ convolution followed by a batch normalization layer is added upon the encoder to produce the log-of-variance $\log\sigma_\theta^2$.

\vspace{2mm}
{\noindent\bf Implementation Details} Both the AET and the AVT networks are trained by the SGD with a batch size of $512$ original images and their transformed versions. Momentum and weight decay are set to $0.9$ and $5\times 10^{-4}$. For the AET, the learning rate is initialized to $0.1$ and scheduled to drop by a factor of $5$ after $240$, $480$, $640$, $800$ and $1,000$ epochs. The network is trained for a total of $1,500$ epochs. The AVT network is trained for $4,500$ epochs, and its learning rate is initialized to $10^{-3}$. Then it is gradually decayed to $10^{-5}$ from $3,000$ epochs after it is increased to $5\times 10^{-3}$ at the epoch $50$.

In the AVT, a single representation is randomly sampled from the encoder $p_\theta(\mathbf z|\mathbf t,\mathbf x)$, which is fed into the decoder $q_\phi(\mathbf t|\mathbf x, \mathbf z)$. To fully exploit the uncertainty of the representations, five samples are drawn and averaged as the representation of an image to train the downstream classifiers. We found averaging randomly sampled representations could outperform only using the mean of the representation.

\vspace{2mm}
{\noindent\bf Applied Transformations} Two types of transformations are considered for model training.
One is the affine transformation. It is a composition of a random rotation with $[-180^\circ, 180^\circ]$, a random translation by $\pm 0.2$ of image height and width in both vertical and horizontal directions, and a random scaling factor of $[0.7, 1.3]$, along with a random shearing of $[-30^\circ,30^\circ]$ degree.
The other is the projective transformation, which is formed by
randomly translating four corners of an image in both horizontal and vertical directions by $\pm 0.125$ of its height and width, after it is randomly scaled by $[0.8, 1.2]$ and rotated by $0^\circ, 90^\circ, 180^\circ,$ or $270^\circ$.

\subsubsection{Results}

\begin{table}
\caption{Comparison between unsupervised feature learning methods on CIFAR-10. The fully supervised NIN and the random Init. + conv have the same three-block NIN architecture, but the first is fully supervised while the second is trained on top of the first two blocks that are randomly initialized and stay frozen during training.}\label{tab01}
\centering
 \begin{tabular}{l|c} \toprule
Method&Error rate\\ \midrule
Supervised NIN \cite{gidaris2018unsupervised} (Upper Bound)&7.20  \\
Random Init. + conv \cite{gidaris2018unsupervised} (Lower Bound)&27.50  \\ \midrule
Roto-Scat + SVM \cite{oyallon2015deep} &17.7 \\
ExamplarCNN \cite{dosovitskiy2014discriminative} &15.7 \\
DCGAN \cite{radford2015unsupervised}&17.2 \\
Scattering \cite{oyallon2017scaling}&15.3\\
RotNet + non-linear \cite{gidaris2018unsupervised}&10.94\\
RotNet + conv \cite{gidaris2018unsupervised}&8.84\\\midrule
AET-affine + non-linear  &9.77\\
AET-affine + conv  &8.05\\
AET-project + non-linear  &9.41\\
AET-project + conv  &7.82\\ \midrule
AVT-project + non-linear &\textbf{8.96} \\
AVT-project + conv &\textbf{7.75} \\
\bottomrule
\end{tabular}
\end{table}

{\noindent\bf Comparison with Other Methods.} To evaluate the effectiveness of a learned unsupervised representation, a classifier is usually trained upon it. In our experiments, we follow the existing evaluation protocols \cite{oyallon2015deep,dosovitskiy2014discriminative,radford2015unsupervised,oyallon2017scaling,gidaris2018unsupervised} by building a classifier on top of the second convolutional block.

First, we evaluate the classification results by using the AET and AVT representations with both model-based and model-free classifiers.  For the model-based classifier, we follow \cite{gidaris2018unsupervised} by training a non-linear classifier with three Fully-Connected (FC) layers -- each of the two hidden layers has $200$ neurons with batch-normalization and ReLU activations, and the output layer is a soft-max layer with ten neurons each for an image class. We also test a convolutional classifier upon the unsupervised features by adding a third NIN block whose output feature map is averaged pooled and connected to a linear soft-max classifier.

Table~\ref{tab01} shows the results by different models. It compares both fully supervised and unsupervised methods on CIFAR-10. The unsupervised AET and AVT with the convolutional classifier almost achieves the same error rate as its fully supervised NIN counterpart with four convolutional blocks ($7.82\%$ and $7.75\%$ vs. $7.2\%$).


\begin{table}
\caption{Error rates of different classifiers on CIFAR 10.}\label{tab02}
\centering
 \begin{tabular}{c|cccc} \toprule
   &1 FC&2 FC&3 FC&conv\\ \midrule
RotNet \cite{gidaris2018unsupervised}&18.21&11.34&10.94 &8.84 \\\midrule
AET-affine  &17.16 &9.77 &10.16 &8.05\\
AET-project &16.65 &9.41 &9.92 &7.82 \\ \midrule
AVT-project &\textbf{16.19} &\textbf{8.96} &\textbf{9.55} &\textbf{7.75}\\\bottomrule
\end{tabular}
\end{table}

We also compare the models when trained with varying number of FC layers in Table~\ref{tab02}. The results show that the AVT leads the AET can consistently achieve the smallest errors no matter which classifiers are used.



\begin{table}[t!]
\caption{The comparison of the KNN error rates by different models with varying numbers $K$ of nearest neighbors on CIFAR-10.}\label{tab04}
\centering
\small
 \begin{tabular}{c|ccccc} \toprule
$K$   &3&5&10&15&20\\ \midrule
RotNet \cite{gidaris2018unsupervised}&25.67&25.01&24.97&25.85&26.00 \\\midrule
AET-affine &24.88&23.29&23.07&23.34&23.94\\
AET-project &23.29&22.40&\bf 22.39&23.32&23.73 \\\midrule
AVT-project &\bf 22.46&\bf 21.62&23.7&\bf 22.16&\bf 21.51 \\ \bottomrule
\end{tabular}
\end{table}

We also note that the probabilistic AVT outperforms the deterministic AET in experiments. This is likely due to the ability of the AVT modeling the uncertainty of representations in training the downstream classifiers. We also find that the projective transformation also performs better than the affine transformation when they are used to train the AET, and thus we mainly use the projective transformation to train the AVT.

{\noindent\bf Comparison based on Model-free KNN Classifiers.}  We also test the model-free KNN classifier based on the averaged-pooled feature representations from the second convolutional block. The KNN classifier is model-free without training a classifier from labeled examples.  This enables us to make a direct evaluation on the quality of learned features.
Table~\ref{tab04} reports the KNN results with varying numbers of nearest neighbors. Again, both the AET and the AVT representations outperform the compared model with varying $K$ nearest neighbors for classification.

{\noindent\bf Comparison with Few Labeled Data.} We also conduct experiments when a small number of labeled examples are used to train the downstream classifiers with the learned representations.
Table~\ref{tab03} reports the results of different models on CIFAR-10. Both the AET and AVT outperform the fully supervised models as well as the other unsupervised models when only few labeled examples ($\leq 1000$ samples per class) are available.

\begin{table*}
\caption{Error rates on CIFAR-10 when different numbers of samples per class are used to train the downstream classifiers. A third convolutional block is trained with the labeled examples on top of the first two NIN blocks of unsupervised representations trained with all unlabeled data. We also compare with the fully supervised models when they are trained with the labeled examples from scratch.}\label{tab03}
\centering
 \begin{tabular}{l|ccccc} \toprule
   &20&100&400&1000&5000\\ \midrule
Supervised conv &66.34&52.74 &25.81 &16.53 &6.93\\
Supervised non-linear &65.03&51.13 &27.17 &16.13 &7.92\\
RotNet + conv \cite{gidaris2018unsupervised}&35.37 &24.72&17.16&13.57 &8.05 \\\midrule
AET-project + conv  &\bf 34.83&24.35 &16.28 &12.58 &7.82 \\
AET-project + non-linear  &\bf 37.13&25.19&18.32 & 14.27 &9.41 \\
\midrule
AVT-project + conv &35.44&\textbf{24.26} &\textbf{15.97} &\textbf{12.27} &\textbf{7.75}\\
AVT-project + non-linear &37.62&\textbf{25.01} &\textbf{17.95} &\textbf{14.14} &\textbf{8.96}\\\bottomrule
\end{tabular}
\end{table*}


\subsection{ImageNet Experiments}

We further evaluate the performance by AET and AVT on the ImageNet dataset.

\subsubsection{Architectures and Training Details}
For a fair comparison with the existing method \cite{noroozi2016unsupervised,bojanowski2017unsupervised,gidaris2018unsupervised}, two AlexNet branches with shared parameters are created with original and transformed images as inputs to train unsupervised models, respectively. The $4,096$-d output features from the second last fully connected layer in each branch are concatenated and fed into the transformation decoder. We still use SGD to train the network, with a batch size of $768$ images and the transformed counterparts, a momentum of $0.9$, a weight decay of $5\times 10^{-4}$.

For the AET model, the initial learning rate is set to $0.01$, and it is dropped by a factor of $10$ at epoch 100 and 150. The model is trained for $200$ epochs in total. For the AVT, the initial learning rate is set to $10^{-3}$, and it is dropped by a factor of $10$ at epoch 300 and 350. The AVT is trained for $400$ epochs in total. We still use the average over five samples from the encoder outputs to train the downstream classifiers to evaluate the AVT. Since the projective transformation has shown better performances, we adopt it for the experiments on ImageNet.

\subsubsection{Results}

\begin{table}
\caption{Top-1 accuracy with non-linear layers on ImageNet. AlexNet is used as backbone to train the unsupervised models. After unsupervised features are learned, nonlinear classifiers are trained on top of Conv4 and Conv5 layers with labeled examples to compare their performances. We also compare with the fully supervised models and random models that give upper and lower bounded performances. For a fair comparison, only a single crop is applied and no dropout or local response normalization is applied during the testing. }\label{tab05}
\centering
 \begin{tabular}{l|cc} \toprule
Method&Conv4 &Conv5\\ \midrule
Supervised from \cite{bojanowski2017unsupervised}(Upper Bound)&59.7&59.7  \\
Random from \cite{noroozi2016unsupervised} (Lower Bound)&27.1 &12.0  \\ \midrule
Tracking \cite{wang2015unsupervised} &38.8&29.8 \\
Context \cite{doersch2015unsupervised} &45.6&30.4 \\
Colorization \cite{zhang2016colorful}&40.7&35.2 \\
Jigsaw Puzzles \cite{noroozi2016unsupervised}&45.3&34.6\\
BIGAN \cite{donahue2016adversarial}&41.9&32.2\\
NAT \cite{bojanowski2017unsupervised}&-&36.0\\
DeepCluster \cite{caron2018deep} &-&44.0\\
RotNet \cite{gidaris2018unsupervised}&50.0&43.8\\\midrule
AET-project  &{53.2}&{47.0}\\
AVT-project &\textbf{54.2}&\textbf{48.4}\\\bottomrule
\end{tabular}
\end{table}

\begin{table*}
\caption{Top-1 accuracy with linear layers on ImageNet. AlexNet is used as backbone to train the unsupervised models under comparison. A $1,000$-way linear classifier is trained upon various convolutional layers of feature maps that are spatially resized to have about $9,000$ elements. Fully supervised and random models are also reported to show the upper and the lower bounds of unsupervised model performances. Only a single crop is used and no dropout or local response normalization is used during testing, except the models denoted with * where ten crops are applied to compare results.}\label{tab06}
\centering
 \begin{tabular}{l|ccccc} \toprule
Method&Conv1 &Conv2&Conv3&Conv4&Conv5\\ \midrule
ImageNet labels(Upper Bound)&19.3&36.3&44.2&48.3&50.5  \\
Random (Lower Bound)&11.6 &17.1&16.9&16.3&14.1  \\
Random rescaled \cite{krahenbuhl2015data}&17.5 &23.0&24.5&23.2&20.6  \\
\midrule
Context \cite{doersch2015unsupervised} &16.2&23.3&30.2&31.7&29.6 \\
Context Encoders \cite{pathak2016context}&14.1&20.7&21.0&19.8&15.5 \\
Colorization\cite{zhang2016colorful}&12.5&24.5&30.4&31.5&30.3\\
Jigsaw Puzzles \cite{noroozi2016unsupervised}&18.2&28.8&34.0&33.9&27.1\\
BIGAN \cite{donahue2016adversarial}&17.7&24.5&31.0&29.9&28.0\\
Split-Brain \cite{zhang2017split}&17.7&29.3&35.4&35.2&32.8\\
Counting \cite{zhang2017split}&18.0&30.6&34.3&32.5&25.7\\
RotNet \cite{gidaris2018unsupervised}&18.8&31.7&38.7&38.2&36.5\\\midrule
AET-project &19.2&32.8&40.6&39.7&37.7\\
AVT-project &\bf 19.5&\bf 33.6&\bf 41.3&\bf 40.3&\bf 39.1\\
\bottomrule
\toprule
DeepCluster* \cite{caron2018deep} &13.4&32.3&41.0&39.6&38.2\\\midrule
AET-project*  &19.3&35.4&44.0&43.6&42.4\\
AVT-project*&\textbf{20.9}&\textbf{36.1}&\textbf{44.4}&\textbf{44.3}&\textbf{43.5}\\\bottomrule
\end{tabular}
\end{table*}

\begin{table*}
\caption{Top-1 accuracy on the Places dataset. A $205$-way logistic regression classifier is trained on top of various layers of feature maps that are spatially resized to have about $9,000$ elements. All unsupervised features are pre-trained on the ImageNet dataset, and then frozen when training the logistic regression classifiers with Places labels. We also compare with fully-supervised networks trained with Places Labels and ImageNet labels, as well as with random models. The highest accuracy values are in bold and the second highest accuracy values are underlined.}\label{tab07}
\centering
 \begin{tabular}{l|ccccc} \toprule
Method&Conv1 &Conv2&Conv3&Conv4&Conv5\\ \midrule
Places labels(Upper Bound)\cite{zhou2014learning}&22.1&35.1&40.2&43.3&44.6 \\
ImageNet labels&22.7&34.8&38.4&39.4&38.7\\
Random (Lower Bound)&15.7 &20.3&19.8&19.1&17.5  \\
Random rescaled \cite{krahenbuhl2015data}&21.4 &26.2&27.1&26.1&24.0  \\
\midrule
Context \cite{doersch2015unsupervised} &19.7&26.7&31.9&32.7&30.9 \\
Context Encoders \cite{pathak2016context}&18.2&23.2&23.4&21.9&18.4 \\
Colorization\cite{zhang2016colorful}&16.0&25.7&29.6&30.3&29.7\\
Jigsaw Puzzles \cite{noroozi2016unsupervised}&\underline{23.0}&31.9&35.0&34.2&29.3\\
BIGAN \cite{donahue2016adversarial}&22.0&28.7&31.8&31.3&29.7\\
Split-Brain \cite{zhang2017split}&21.3&30.7&34.0&34.1&32.5\\
Counting \cite{zhang2017split}&\textbf{23.3}&\textbf{33.9}&36.3&34.7&29.6\\
RotNet \cite{gidaris2018unsupervised}&21.5&31.0&35.1&34.6&33.7\\\midrule
 AET-project & 22.1&32.9&\underline{37.1}&\underline{36.2}&\underline{34.7}\\
AVT-project &22.3&\underline{33.1}&\textbf{37.8}&\textbf{36.7}&\textbf{35.6}\\\bottomrule
\end{tabular}
\end{table*}

Table~\ref{tab05} reports the Top-1 accuracies of the compared methods on ImageNet by following the evaluation protocol in \cite{noroozi2016unsupervised}. Two settings are adopted for evaluation, where Conv4 and Conv5 mean to train the remaining part of AlexNet on top of Conv4 and Conv5 with the labeled data. All the bottom convolutional layers up to Conv4 and Conv5 are frozen after they are trained in an unsupervised fashion. 
From the results, in both settings, the AVT model consistently outperforms the other unsupervised models, including the AET.

We also compare with the fully supervised models that give the upper bound of the classification performance by training the AlexNet with all labeled data end-to-end. The classifiers of random models are trained on top of Conv4 and Conv5 whose weights are randomly sampled, which set the lower bounded performance. By comparison, the AET models narrow the performance gap to the upper bound supervised models from $9.7\%$ and $15.7\%$ by RotNet and DeepCluster on Conv4 and Conv5, to $6.5\%$ and $12.7\%$ by the AET, and to $5.5\%$ and $11.3\%$ by the AVT.


Moreover, we also follow the testing protocol adopted in \cite{zhang2017split} to compare the models by training a $1,000$-way linear classifier on top of different numbers of convolutional layers in Table~\ref{tab06}.  Again, the AVT consistently outperforms all the compared unsupervised models in terms of the Top-1 accuracy.

\subsection{Places Experiments}

We also compare different models on the Places dataset. Table~\ref{tab07} reports the results. Unsupervised models are pretrained on the ImageNet dataset, and a linear logistic regression classifier is trained on top of different layers of convolutional feature maps with Places labels. It assesses the generalizability of unsupervised features from one dataset to another. The models are still based on AlexNet variants. We compare with the fully supervised models trained with the Places labels and ImageNet labels respectively, as well as with the random networks. Both the AET and the AVT models outperform the other unsupervised models, except performing slightly worse than Counting \cite{zhang2017split} with a shallow representation by Conv1 and Conv2.

\section{Experiments: (Semi-)Supervised Learning}\label{sec:supexp}

We compare the proposed SAT model with the other state-of-the-art semi-supervised methods in this section. For the sake of fair comparison, we follow the test protocol used in literature \cite{tarvainen2017mean,laine2016temporal} on both CIFAR-10 \cite{krizhevsky2009learning} and SVHN \cite{netzer2011reading}, which are widely used as the benchmark datasets to evaluate the semi-supervised models.

%

\subsection{Network Architecture and Implementation Details}

{\bf\noindent Network Architecture} For the sake of a fair comparison, a 13-layer convolutional neural network, which has been widely used in existing semi-supervised models \cite{laine2016temporal,tarvainen2017mean,miyato2018virtual}, is adopted as the backbone to build the SAT. 
It consists of three convolutional blocks, each of which contains three convolution layers. The SAT has two branches of such three blocks with shared weights, each taking the original and transformed images as input, respectively. The output feature maps from the third blocks of two branches are concatenated and average-pooled, resulting in a $256$-d feature vector. A fully-connected layer follows to predict the mean $d_\phi$ and the log-of-variance $\log \sigma_\phi^2$ of the transformation.
The first two blocks are used as the encoder to output the mean $f_\theta$ of the representation, upon which an additional $1\times 1$ convolution layer with batch normalization is added to compute the log-of-variance $\log\sigma_\theta^2$.

In addition, a classifier head is built on the representation from the encoder. Specifically, we draw five random representations of an input image, and feed their average to the classifier. The classifier head has the same structure as the third convolutional block but its weights differ from the Siamese branches of transformation decoder. The output feature map of this convolutional block is globally average-pooled to $128$-d feature vector, and a softmax fully connected layer follows to predict the image label.

{\bf\noindent Implementation Details} The representation encoder, transformation decoder and the classifier are trained in an end-to-end fashion.
In particular, the SGD is adopted to iteratively update their weights over a minbatch with $500$ images, their transformed counterparts, and $40$ labeled examples. Momentum and weight decay are set to $0.9$ and $5\times 10^{-4}$, respectively. The model is trained for a total of $4,500$ epochs. The learning rate is initialized to $10^{-3}$. It is increased to $5\times 10^{-3}$ at epoch $50$, before it is linearly decayed to $10^{-5}$ starting from $3,000$ epochs.  For a fair comparison, we adopt the entropy minimization used in the state-of-the-art virtual adversarial training \cite{miyato2018virtual}. A standard set of data augmentations in literature \cite{laine2016temporal,tarvainen2017mean,miyato2018virtual} are also adopted through experiments, which include both horizontal flips and random translations on CIFAR-10, and only random translations on SVHN. The projective transformation that performs the better than the affine transformation is adopted to train the semi-supervised representations.


\subsection{Results}

\begin{table*}
\caption{Error rate percentage of compared methods on CIFAR-10 over ten runs (four runs when all labels are used). }\label{tab:cifar10}
\vspace{2mm}
\centering
 \begin{tabular}{ccccc} \toprule
   &1000 labels&2000 labels&4000 labels&50000 labels\\ \midrule
GAN \cite{salimans2016improved}&&&18.63$\pm$ 2.32 \\
$\Pi$ model \cite{laine2016temporal}&&&12.36$\pm$ 0.31 &5.56$\pm$0.10  \\ \
Temporal Ensembling \cite{laine2016temporal}&&&12.16$\pm$ 0.31 &5.60$\pm$0.10\\
VAT \cite{miyato2018virtual} &&&10.55\\
Supervised-only&46.43$\pm$1.21&33.94$\pm 0.73$&20.66$\pm$0.57&5.81$\pm$0.15\\
$\Pi$ model \cite{tarvainen2017mean} &27.36$\pm$1.20&18.02$\pm$0.60&13.20$\pm$0.27&6.06$\pm$0.11\\
Mean Teacher \cite{tarvainen2017mean} &21.55$\pm$1.48&15.73$\pm$0.31&12.31$\pm$0.28&5.94$\pm$0.15\\\midrule
SAT&{\bf 14.89$\pm$0.38}&{\bf 11.71$\pm$0.29} &\textbf{9.58$\pm$0.11}&{\bf 4.91$\pm$0.13}\\
\bottomrule
\end{tabular}
\end{table*}

\begin{table*}
\caption{Error rate percentage of compared methods on SVHN over ten runs (four runs when all labels are used).}\label{tab:svhn}
\vspace{2mm}
\centering
 \begin{tabular}{cccccc} \toprule
   &250 labels&500 labels&1000 labels&73257 labels\\ \midrule
GAN \cite{salimans2016improved}&&18.44$\pm$4.8&8.11$\pm$ 11.3 \\
$\Pi$ model \cite{laine2016temporal}&&6.65$\pm$0.53&4.82$\pm$ 0.17&2.54$\pm$0.04  \\ \
Temporal Ensembling \cite{laine2016temporal}&&5.12$\pm$0.13&4.42$\pm$ 0.16 &2.74$\pm$0.06\\
VAT \cite{miyato2018virtual} &&&3.86\\
Supervised-only&27.77$\pm$3.18&16.88$\pm 1.30$&12.32$\pm$0.95&2.75$\pm$0.10\\
$\Pi$ model \cite{tarvainen2017mean}&9.69$\pm$0.92&6.83$\pm$0.66&4.95$\pm$0.26&2.50$\pm$0.07\\
Mean Teacher \cite{tarvainen2017mean}&4.35$\pm$0.50&4.18$\pm$0.27&3.95$\pm$0.19&2.50$\pm$0.05\\\midrule
SAT&\textbf{4.30$\pm$0.22}&\textbf{3.72$\pm$0.20}&\textbf{3.44$\pm$0.10}&\textbf{2.15$\pm$0.06}\\
\bottomrule
\end{tabular}
\end{table*}

We compare with the state-of-the-art semi-supervised methods in literature \cite{tarvainen2017mean,laine2016temporal}.
Table~\ref{tab:cifar10} and \ref{tab:svhn} show that the SAT outperforms the compared methods with different numbers of labeled examples on both CIFAR-10 and SVHN datasets.
The results demonstrate that the SAT has captured the useful representation, which delivers competitive classification performances from the transformations on both unlabeled and labeled examples to semi-supervise the network training with only few labeled examples.

In particular, the proposed SAT reduces the average error rates of Mean Teacher (the second best performing method) by 30.9\%, 25.6\%, 22.2\% relatively with $1,000$, $2,000$, and $4,000$ labels on CIFAR-10, while reducing them by $1.1\%$, $11\%$, $12.9\%$ relatively with $250$, $500$, and $1,000$ labels on SVHN.
The compared semi-supervised methods, including $\Pi$ model \cite{laine2016temporal}, Temporal Ensembling \cite{laine2016temporal}, and Mean Teacher \cite{tarvainen2017mean}, attempt to maximize the consistency of model predictions on the transformed and original images to train semi-supervised classifiers. While they also apply the transformations to explore unlabeled examples, the competitive performance of the SAT model shows
the transformation-equivariant representations are more compelling for classifying images than those compared methods predicting consistent labels under transformations. It justifies the proposed criterion of pursuing the transformation equivariance as a regularizer to train a classifier.


\begin{table*}[t!]
\caption{Comparison of error rate percentages of SAT and VAT with and without Entropy Minimization (EntMin) on CIFAR-10.}\label{tab:ablation}
\centering
 \begin{tabular}{ccccc} \toprule
   &1000 labels&2000 labels&4000 labels&50000 labels\\ \midrule
VAT w/o EntMin\cite{miyato2018virtual} &&&11.36\\
SAT w/o EntMin&15.32$\pm$0.40&12.76$\pm$0.26&10.90$\pm$0.21&5.95$\pm$0.17\\\midrule
VAT with EntMin\cite{miyato2018virtual} &&&10.55\\
SAT with EntMin&\textbf{14.89$\pm$0.38}&\textbf{11.71$\pm$0.29}&\textbf{9.58$\pm$0.11}&\textbf{4.91$\pm$0.13}\\
\bottomrule
\end{tabular}
\end{table*}

It is not hard to see that the SAT can be integrated into the other semi-supervised methods as their base representations, and we believe this could further boost their performances. This will be left to the future work as it is beyond the scope of this paper.



\subsubsection{The Impact of Entropy Minimization}

We also conduct an ablation study of the Entropy Minimization (EntMin) on the model performance. EntMin was used in VAT \cite{miyato2018virtual} that outperformed the other semi-supervised methods in literature. Here, we compare the error rates between the SAT and the VAT with or without the EntMin.
As shown in Table~\ref{tab:ablation}, no matter if the entropy minimization is adopted, the SAT always outperforms the corresponding VAT. We also note that, even without entropy minimization, the SAT still performs better than the other state-of-the-art semi-supervised classifiers such as Mean Teacher, Temporal Ensembling, and $\Pi$ model shown in Table~\ref{tab:cifar10}. This demonstrates the compelling performance of the SAT model.

\subsubsection{Comparison with Data Augmentation by Transformations}

\begin{table}
\caption{Error rate percentage of Data Augmentation by Transformations (DAT) on CIFAR-10. To ensure a fair comparison, the same set of labeled examples are split from the training set for semi-supervised learning.}\label{tab:da}
\vspace{2mm}\centering
 \begin{tabular}{ccccc} \toprule
   &1000 labels&2000 labels&4000 labels\\ \midrule
DAT&51.00&38.61&27.99\\
SAT&\textbf{15.72}&\textbf{13.20}&\textbf{11.05}\\\bottomrule
\end{tabular}
\vspace{-10pt}
\end{table}

We also compare the performances between the SAT and a classification network trained with the augmented images by the transformations.
Specifically, in each minibatch, input images are augmented with the same set of random projective transformations used in the SAT. The transformation-augmented images and their labels are used to train a network with the same 13-layer architecture that has been adopted as the SAT backbone. Note that the transformation augmentations are applied on top of the standard augmentations mentioned in the implementation details for a fair comparison with the SAT.


Table~\ref{tab:da} compares the results between the SAT and the Data Augmentation by Transformation (DAT) classifier on CIFAR-10. It shows the SAT significantly outperforms the DAT. This is not surprising -- data augmentation by transformations can only augment the labeled examples, limiting its ability of exploring unlabeled examples that play very important roles in semi-supervised learning.


Moreover, the projective transformations used in the SAT could severely distort training images that could incur undesired update to the model weights if the distorted images were used to naively train the network. This is witnessed by the results that the data augmentation by transformations performs even worse than the supervised-only method (see Table~\ref{tab:cifar10}).


In contrast, the SAT avoids a direct use of the transformed images to supervise the model training with their labels. Instead, it trains the learned representations to contain as much information as possible about the transformations. The superior performance demonstrates its outstanding ability of classifying images by exploring the variations of visual structures induced by transformations on both labeled and unlabeled images.



\section{Conclusion and Future Works}\label{sec:concl}
In this paper, we present to use a novel approach of AutoEncoding Transformations (AET) to learn representations that equivary to applied transformations on images. Unlike the group equivariant convolutions that would become intractable with a composition of complex transformations, the AET model seeks to learn representations of arbitrary forms by reconstructing transformations from the encoded representations of original and transformed images. The idea is further extended to a probabilistic model by  maximizing the mutual information between the learned representation and the applied transformation. The intractable maximization problem is handled by introducing a surrogate transformation decoder and maximizing a variational lower bound of the mutual information, resulting in the Autoencoding Variational Transformations (AVT). Along this direction, a (Semi-)Supervised Autoencoding Transformation (SAT) approach can be derived by maximizing the joint mutual information of the learned representation with both the transformation and the label for a given sample. The proposed AET paradigm lies a solid foundation to explore transformation equivariant representations in many learning tasks. Particularly, we conduct experiments to show its superior performances on both unsupervised learning to semi-(supervised) learning tasks following standard evaluation protocols. In future, we will explore the great potential of applying the learned AET representation as the building block on more learning tasks, such as (instance) semantic segmentation, object detection, super-resolution reconstruction, few-shot learning, and fine-grained classification.


%




\ifCLASSOPTIONcaptionsoff
  \newpage
\fi



\bibliographystyle{IEEEtran}
\bibliography{egbib,aet_egbib,small_data}

%
\vspace{-12mm}
\begin{IEEEbiography}[{\includegraphics[width=1in,height=1.25in,clip,keepaspectratio]{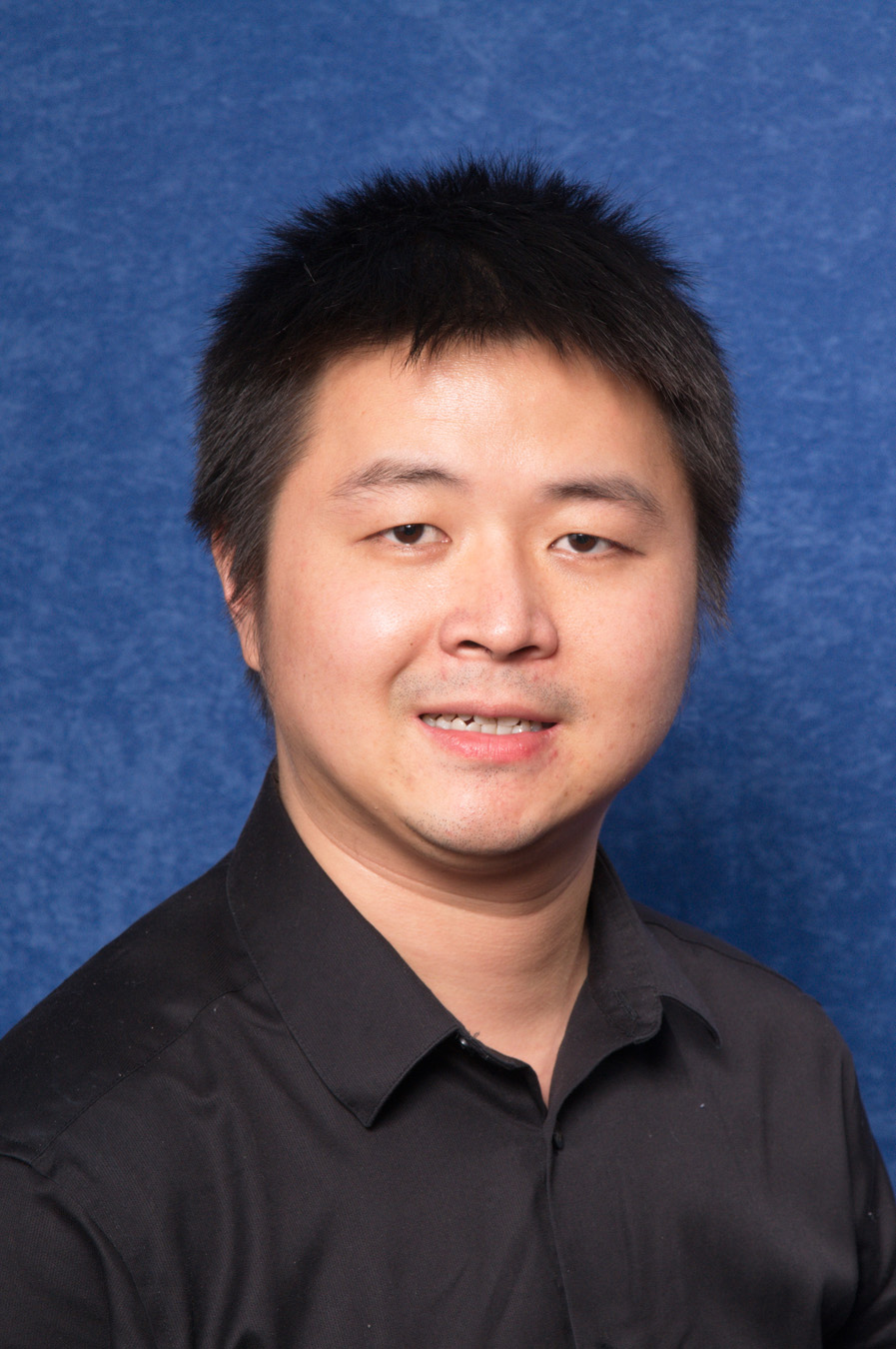}}]{Guo-Jun Qi}
Guo-Jun Qi (M14-SM18) is the Chief Scientist leading and overseeing an international R\&D team for multiple artificial intelligent services on the Huawei Cloud since August 2018. He was a faculty member in the Department of Computer Science and the director of MAchine Perception and LEarning (MAPLE) Lab at the University of Central Florida since August 2014. Prior to that, he was also a Research Staff Member at IBM T.J. Watson Research Center, Yorktown Heights, NY.
His research interests include machine learning and knowledge discovery from multi-modal data sources to build smart and reliable information and decision-making systems.
Dr. Qi has published more than 100 papers in a broad range of venues in pattern recognition, machine learning and computer vision.
He also has served or will serve as a general co-chair for ICME 2021, technical program co-chair for ACM Multimedia 2020, ICIMCS 2018 and MMM 2016, as well as an area chair (senior program committee member) for multiple academic conferences. Dr. Qi is an associate editor for IEEE Transactions on Circuits and Systems for Video Technology (T-CSVT), IEEE Transactions on Multimedia (T-MM), IEEE Transactions on Image Processing (T-IP), Pattern Recognition (PR), and ACM Transactions on Knowledge Discovery from Data (T-KDD).
\end{IEEEbiography}

\vspace{-12mm}
\begin{IEEEbiography}[{\includegraphics[width=1in,height=1.25in,clip,keepaspectratio]{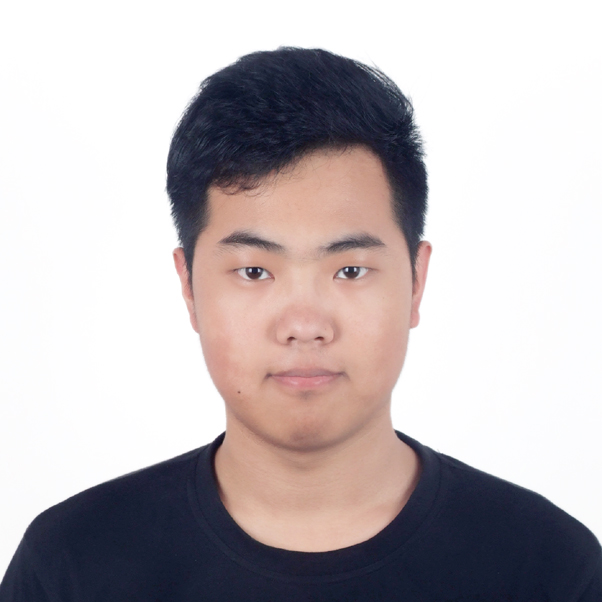}}]{Liheng Zhang}
Liheng Zhang received the B. S. degree in School of Electronic Information and Communications in Huazhong University of Science and Technology, in 2015. He is currently pursuing the PhD degree with the Department of Computer Science, University of Central Florida. His research interests include deep learning, machine learning and computer vision.
\end{IEEEbiography}

\vspace{-12mm}
\begin{IEEEbiography}[{\includegraphics[width=1in,height=1.25in,clip,keepaspectratio]{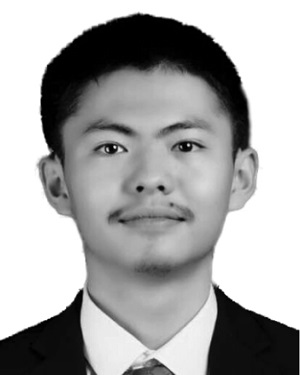}}]{Xiao Wang}
Xiao Wang received the B.S. degree in Department of Computer Science in Xi'an Jiaotong University, in 2018. He is currently pursuing the PhD degree with the Department of Computer Science, Purdue University.  His research interests include deep learning, computer vision, bioinformatics and intelligent systems.
\end{IEEEbiography}



\end{document}